\documentclass[letterpaper]{article}
\usepackage{aaai2026}
\usepackage{times}
\usepackage{helvet}
\usepackage{courier}
\usepackage[hyphens]{url}
\usepackage{graphicx}
\usepackage{natbib}
\usepackage{caption}
\usepackage{algorithm}
\usepackage{algorithmic}
\usepackage{amsmath}
\usepackage{booktabs}
\usepackage{multirow}
\usepackage{amssymb}
\usepackage{pifont}
\usepackage{makecell}
\usepackage{adjustbox}

\usepackage{newfloat}
\usepackage{listings}
\DeclareCaptionStyle{ruled}{labelfont=normalfont,labelsep=colon,strut=off}
\floatstyle{ruled}
\newfloat{listing}{tb}{lst}{}
\floatname{listing}{Listing}

\title{STELLAR: Scene Text Editor for Low-Resource Languages and Real-World Data}
\author{
    Yongdeuk Seo\textsuperscript{\rm 1},
    Hyun-seok Min\textsuperscript{\rm 2}, 
    Sungchul Choi\textsuperscript{\rm 1}\thanks{Corresponding author.}
}
\affiliations{
    \textsuperscript{\rm 1}Major in Industrial Data Science \& Engineering, Department of Industrial and Data Engineering, Pukyong National University, \\
    \textsuperscript{\rm 2}Tomocube Inc. \\

    sod7050@pukyong.ac.kr,
    hsmin@tomocube.com,
    sc82.choi@pknu.ac.kr

}

\begin{document}

\maketitle

\begin{abstract}
Scene Text Editing (STE) is the task of modifying text content in an image while preserving its visual style, such as font, color, and background. While recent diffusion-based approaches have shown improvements in visual quality, key limitations remain: lack of support for low-resource languages, domain gap between synthetic and real data, and the absence of appropriate metrics for evaluating text style preservation. To address these challenges, we propose \textbf{STELLAR} (\textbf{S}cene \textbf{T}ext \textbf{E}ditor for \textbf{L}ow-resource \textbf{LA}nguages and \textbf{R}eal-world data). STELLAR enables reliable multilingual editing through a language-adaptive glyph encoder and a multi‑stage training strategy that first pre‑trains on synthetic data and then fine-tunes on real images. We also construct a new dataset, STIPLAR(Scene Text Image Pairs of Low-resource lAnguages and Real-world data), for training and evaluation. Furthermore, we propose Text Appearance Similarity (TAS), a novel metric that assesses style preservation by independently measuring font, color, and background similarity, enabling robust evaluation even without ground truth. Experimental results demonstrate that STELLAR outperforms state-of-the-art models in visual consistency and recognition accuracy, achieving an average TAS improvement of 2.2\% across languages over the baselines.
\end{abstract}

\begin{links}
    \link{Code}{github.com/yongchoooon/stellar}
    \link{Datasets}{huggingface.co/datasets/yongchoooon/stiplar}
\end{links}

\section{Introduction}

With the rapid expansion of the global content industry, there is a growing demand for accurate text modification within images in multiple languages. Applications such as advertisement banners, product packaging, game and film localization, and augmented reality signage frequently require changing only the textual part of an image while preserving its original visual style and background. This capability is essential for scalable content production, enabling the reuse and adaptation of visual assets across diverse linguistic and cultural contexts.

\begin{figure}[t]
\centering
\includegraphics[width=0.93\columnwidth]{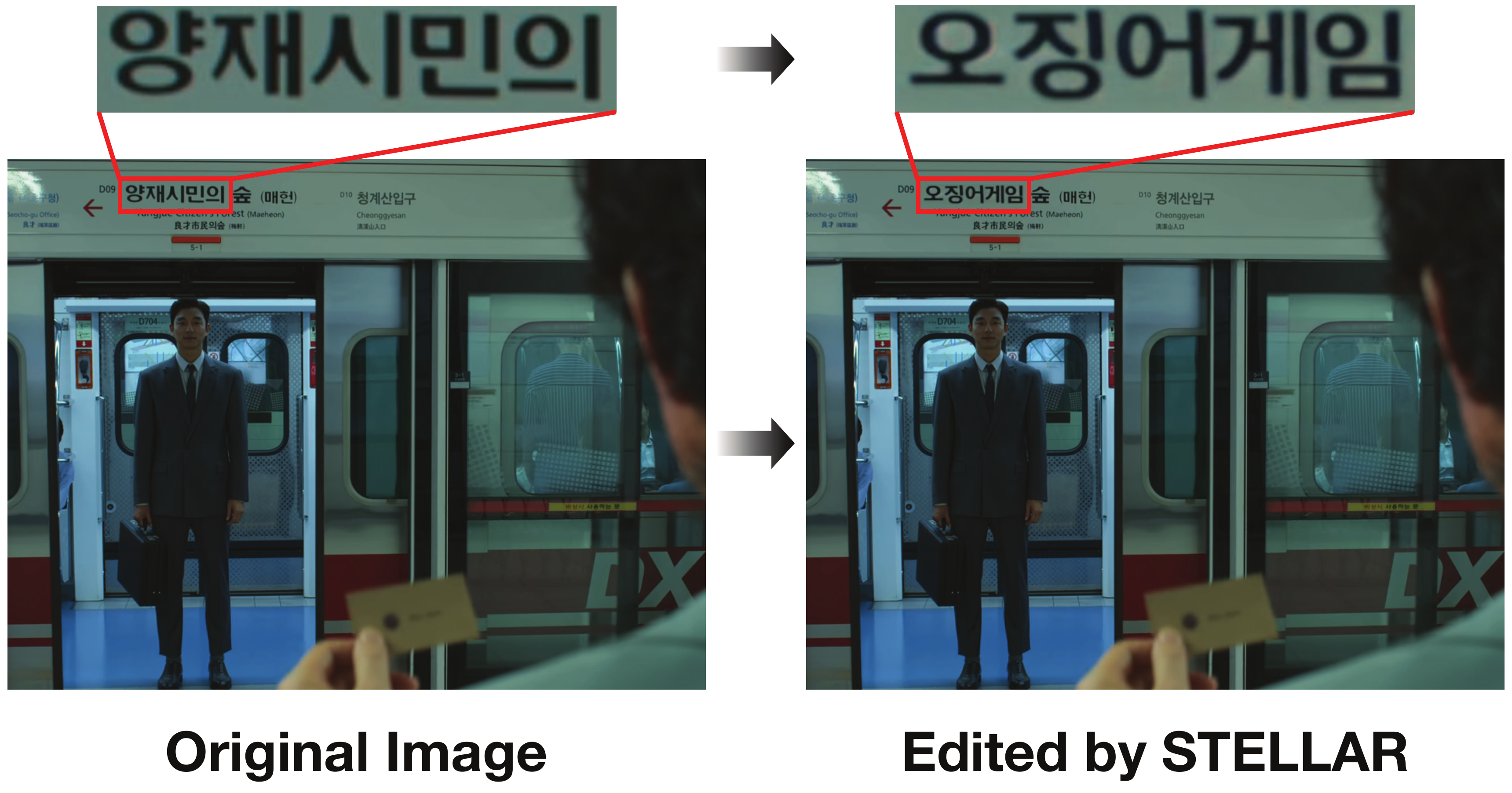}
\caption{Example of text editing performed by STELLAR, showing visually consistent modification of Korean text within a real-world scene.}
\label{fig:main_squidgame}
\end{figure}

The growing presence of multilingual media, including K-culture, Japanese pop content, and Arabic entertainment, further emphasizes the need for high-quality Scene Text Editing (STE) that supports a wide range of scripts and writing systems. Figure~\ref{fig:main_squidgame} presents an illustrative example in which our model edits Korean text within a complex scene, demonstrating its ability to perform visually consistent text modification in real-world conditions.

STE aims to modify text content in images while preserving the original font, color, and background. It has emerged as a key technology to meet these industrial demands~\cite{srnet}. Due to its wide applicability across computer vision, graphics, and design, STE continues to attract increasing research attention~\cite{translatar, mar, textstylebrush}.

Early works on STE were primarily based on Generative Adversarial Networks (GANs)~\cite{gan}, but suffered from instability and limited visual quality. More recent studies employ mask-and-inpaint paradigms powered by Diffusion Models (DMs)~\cite{ldm}, which mask the text region and inpaint it with new content, improving background preservation and image consistency~\cite{glyphdraw, glyphcontrol, diffute, textdiffuser, showmetheworldinmylanguage, udifftext, anytext, anytext2, textgen, brushyourtext, diffste, textdiffuser2, glyphmastero, textflux}. However, these models still struggle to reproduce fine-grained styles such as font and color. Meanwhile, a recent method~\cite{textctrl} adopts direct substitution approaches that disentangle visual style and character structure for explicit conditioning, achieving better style preservation and text rendering quality than mask-and-inpaint models.

Despite recent progress, existing STE methods share three major limitations:

\textbf{(1) Limited support for low-resource languages.} While some recent works~\cite{anytext, anytext2, textflux} explore multilingual scene text generation, they often fail to achieve satisfactory performance for languages with complex scripts or limited data, such as Korean, Arabic, and Japanese. Without language-aware modeling that accounts for diverse character structures, accurate editing and style preservation remain difficult.

\textbf{(2) Domain gap between synthetic and real data.} Most STE models are trained solely on synthetic image pairs~\cite{srnet, swaptext, udifftext, diffste, textctrl}, which fail to capture real-world characteristics, such as lighting, texture, and noise. This domain gap leads to performance degradation during inference, including color distortion and texture artifacts~\cite{mostel, darling, textctrl}.

\textbf{(3) Inadequate evaluation metrics.} Existing studies primarily rely on indirect image similarity metrics such as SSIM, PSNR, MSE, and FID~\cite{fid}, which are inherently unsuitable for evaluating STE performance. These metrics tend to penalize textual content changes when visual styles are preserved and are inapplicable when ground truth is unavailable, limiting real-world applicability.

First, we introduce a language-adaptive glyph encoder pre-trained on synthetic text image pairs for low-resource languages, including Korean, Arabic, and Japanese. This module extracts language-specific glyph structural features, guiding the model to capture visual characteristics essential for accurate text rendering.

Next, considering the scarcity of real-world text image pair datasets for low-resource languages, we collected and curated \textbf{STIPLAR} (\textbf{S}cene \textbf{T}ext \textbf{I}mage \textbf{P}airs of \textbf{L}ow-resource l\textbf{\text{A}}nguages and \textbf{R}eal-world data) dataset from open-source resources and web crawling. By fine-tuning the DM~\cite{ldm} with a small number of real images, we effectively bridge the domain gap between training and inference, enabling realistic text editing in complex real-world scenes. This process enhances visual realism and model applicability to real-world content production pipelines, and the resulting dataset will be publicly released.

Finally, to address the inadequacy of existing evaluation metrics in reflecting text style preservation, we propose Text Appearance Similarity (TAS), a novel metric that independently analyzes similarity for visual attributes such as text color, font, and background. TAS enables quantitative evaluation of visual style similarity independent of text content and provides improved interpretability and practicality as it remains applicable even without ground truth images.

In summary, we propose a robust STE model that achieves effective performance across diverse languages and real-world images through: (1) adapting to low-resource languages via language-aware modeling, (2) constructing and leveraging a real-world text image pair dataset for domain adaptation, and (3) introducing TAS, a new metric for quantitative evaluation of text style similarity. These contributions enhance the practical utility of STELLAR for real-world multilingual content creation and editing workflows.

\section{Related Work}

\subsection{STE Approaches}

STE aims to modify only the text content in an image while preserving its visual style and the background. Early studies primarily focused on GAN-based approaches~\cite{gan} that replaced text regions and reconstructed backgrounds~\cite{srnet, mostel, faster, stefann}, but suffered from training instability and resolution degradation.

With the advent of DMs~\cite{ldm}, mask-and-inpaint based methods have become mainstream\cite{glyphdraw, glyphcontrol, diffute, textdiffuser, showmetheworldinmylanguage, udifftext, anytext, anytext2, textgen, brushyourtext, diffste, textdiffuser2, glyphmastero, textflux}, enabling stable reproduction of image texture and structure by inpainting masked text regions. This approach naturally preserves complex backgrounds and maintains visual consistency, leading many studies to focus on text image generation and demonstrating strong potential for extension to STE. However, it still struggles to accurately reproduce fine-grained text styles such as font and color.

Recent studies~\cite{textctrl, darling} have proposed direct substitution frameworks that disentangle style and content features. These models separate visual style (font, color, background) from text image, and use each as conditional input to generate new text that aligns with the original appearance. Leveraging transformer~\cite{transformer} or DM-based architectures, they enhance both style preservation and text rendering accuracy.

\begin{table}[t]
\begin{adjustbox}{width=\columnwidth, center}
\setlength{\tabcolsep}{2pt}
\centering
\small
\begin{tabular}{l|c|ccccc|c}
\toprule
Dataset & type & Chinese & English & Korean & Arabic & Japanese & Total \\
\midrule
AnyWord-3M   & T2I & 1.6M & 1.39M & 2K   & 2K   & 2K   & 3M \\
TG-2M        & T2I & 1.23M & 1.3M & --   & --   & --   & 2.53M \\
\textbf{STIPLAR}      & I2I & --   & --   & 9.7K & 6.3K & 2K   & \textbf{18K} \\
\bottomrule
\end{tabular}
\end{adjustbox}
\caption{Comparison of multilingual text image datasets~\cite{anytext, textgen}. Type indicates the data pair type.}
\label{tab:dataset_image_count}
\end{table}

\subsection{Multi-Lingual Text Image Generation}

Efforts to realize STE in multilingual contexts are steadily increasing~\cite{anytext, anytext2, textgen, textflux, textmastero}. For instance, AnyWord-3M~\cite{anytext} is a multilingual dataset that mainly comprises English and Chinese text-image pairs but also includes low-resource languages such as Korean, Arabic, Japanese, Hindi, and Bangla. It was curated from diverse sources, including publicly available datasets such as LAION-400M~\cite{laion400m} and MLT-2019~\cite{icdarmlt2019}, and was used to train a multilingual generation framework based on ControlNet~\cite{controlnet}. It has subsequently been employed by several multilingual text image generation and editing studies. TG-2M~\cite{textgen}, primarily comprises English and Chinese text images sourced from multiple open datasets, with captions refined using a vision-language model (VLM)~\cite{blip2, qwenvl} to support multilingual text image generation.

However, as shown in Table~\ref{tab:dataset_image_count}, these datasets predominantly feature English and Chinese content, while the representation of low-resource languages, including Korean, Arabic, and Japanese, remains limited to a few thousand samples, perpetuating linguistic biases. To address this limitation, we construct a new, diverse multilingual scene text image pair dataset and train STE model on it.

\subsection{Domain Adaptation for Real-world data}

STE models are commonly trained on synthetic text image pairs generated by rendering engines~\cite{synthtext, synthtiger}. Although synthetic data enable large-scale and controlled generation, they often fail to capture complex real-world attributes such as lighting, texture, and noise, creating a domain gap that degrades performance during inference on real-world data.

Several studies have acknowledged this domain gap as a critical limitation. MOSTEL~\cite{mostel} noted significant performance declines in real-world scenarios and attempted to mitigate this issue through a style-oriented self-training approach, although it did not fundamentally resolve the domain gap. DARLING~\cite{darling} also recognized the synthetic-to-real domain gap, highlighting the difficulty of leveraging unlabeled real-world data for pre-training and underscoring the need for self-supervised learning methods. TextCtrl~\cite{textctrl} introduced a post-hoc technique to alleviate the domain gap during inference by dynamically incorporating source image information, thereby enhancing style consistency.

\begin{figure}[t]
\centering
\includegraphics[width=\columnwidth]{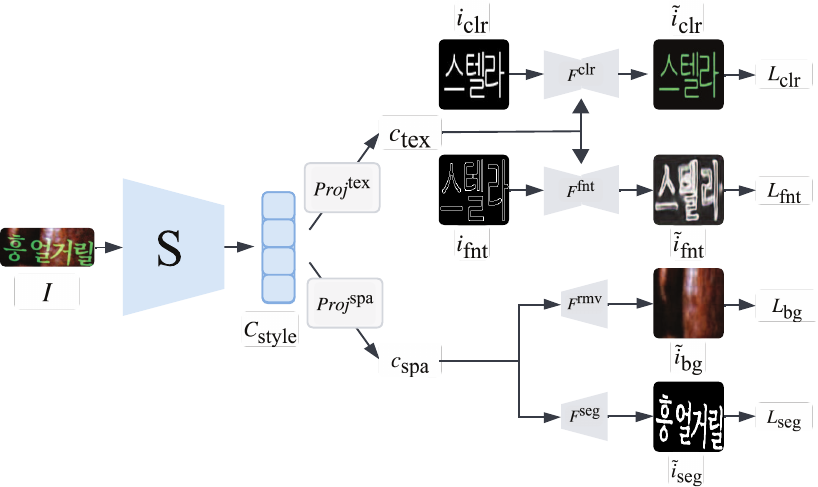}
\caption{Multi-task training scheme of the text style encoder $S$ in baseline.}
\label{fig:text_encoder}
\end{figure}

Most existing studies primarily rely on synthetic data and use real-world images only for evaluation or limited adjustments. In contrast, we directly collect real-world scene text image pairs that include low-resource languages and apply a multi-stage training strategy during training, which yields robust STE across domains without any additional post-hoc technique. 

\begin{figure*}[t]
\centering
\includegraphics[width=0.8\textwidth]{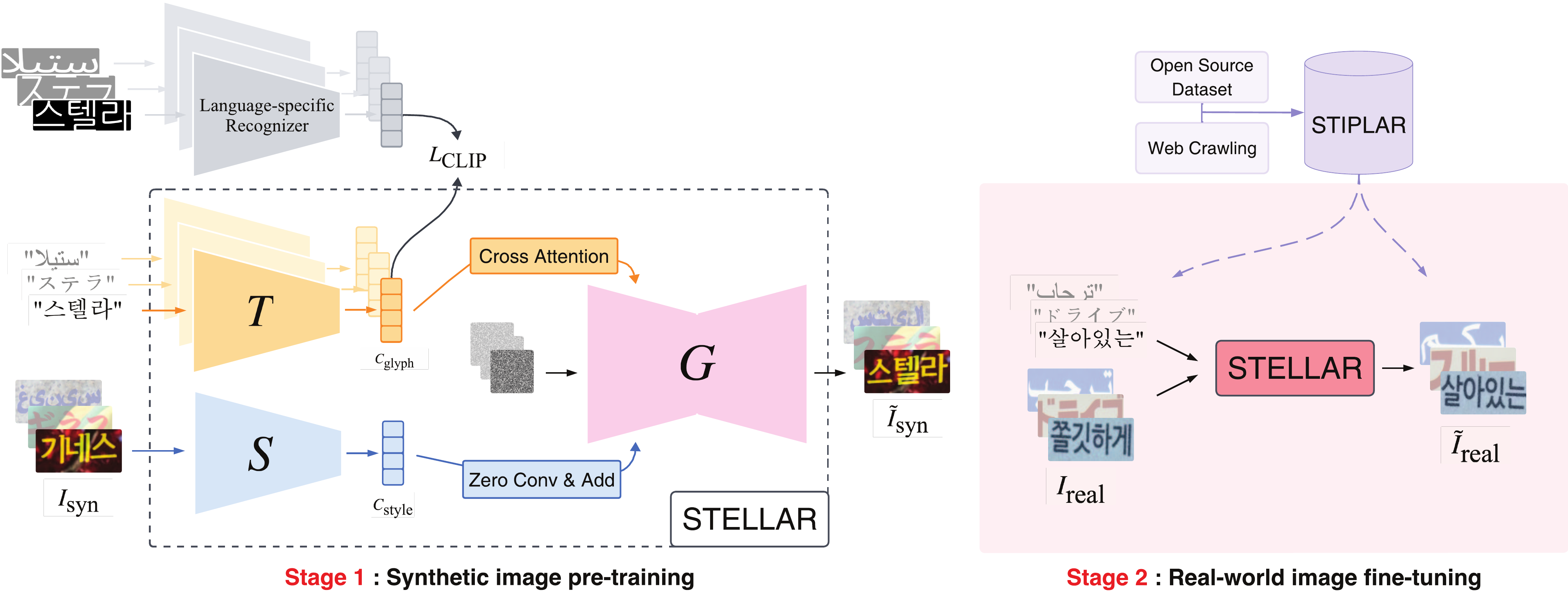} 
\caption{STELLAR framework and training pipeline. Language-adaptive glyph encoder $T$, pre-trained with a language-specific recognizer, extracts glyph features $C_\text{glyph}$ to guide the diffusion generator $G$ via cross-attention. Style features $C_\text{style}$ from pre-trained text style encoder $S$ are injected through skip-connections. $G$ is first pre-trained on synthetic data (Stage 1) and then fine-tuned on real-world images from the STIPLAR dataset (Stage 2).}
\label{fig:main}
\end{figure*}

\section{Method}

\subsection{Preliminary}

We propose a STE framework for low-resource languages that is robust to real-world images, built upon a DM-based baseline~\cite{textctrl}. This baseline separately encodes visual style from the original text image and textual structure from the target text, using these features as conditions to directly substitute new text content.

The architecture consists of two encoders and a diffusion generator, each trained independently. Text glyph structure encoder extracts character-level glyph features $C_{\text{struct}}$ from the target text and aligns them with visual features of rendered text images using CLIP~\cite{clip} Loss. The visual features are obtained from a frozen pre-trained scene text recognizer~\cite{abinet}. During training, each character is rendered with diverse fonts sampled from a large font cluster, inducing variation in glyph structure and enhancing the alignment between textual and visual features.

Text style encoder disentangles fine-grained style components, such as color, font, background, and spatial mask, by leveraging a multi-task learning strategy. The overall training scheme of the encoder is illustrated in Figure~\ref{fig:text_encoder}. Given a text image $I$,  a ViT-based encoder~\cite{vit} extracts style features $C_{\text{style}}$, which is projected into texture features $c_{\text{tex}}$ and a spatial features $c_{\text{spa}}$ using two separate linear projection layers, denoted as $Proj^{\text{tex}}$ and $Proj^{\text{spa}}$, respectively.

\begin{equation}
c_{\text{tex}} = Proj^{\text{tex}}(C_{\text{style}}), c_{\text{spa}} = Proj^{\text{spa}}(C_{\text{style}})
\label{eq:preliminary_tex_spa}
\end{equation}

The texture features $c_{\text{tex}}$ and spatial features $c_{\text{spa}}$ are used for:

\noindent\textbf{Text color transfer.}
A grayscale text image $i_{\text{clr}}$ is colorized to produce the output $\tilde{i}_{\text{clr}}$ using a lightweight encoder-decoder $F^{\text{clr}}$.
\begin{equation}
\tilde{i}_{\text{clr}} = F^{\text{clr}}(c_{\text{tex}}, i_{\text{clr}}).
\label{eq:preliminary_color}
\end{equation}
\noindent\textbf{Text font transfer.}
A template font image $i_{\text{fnt}}$ is transformed to produce the output $\tilde{i}_{\text{fnt}}$ in the style of the source font using another lightweight encoder-decoder $F^{\text{fnt}}$.
\begin{equation}
\tilde{i}_{\text{fnt}} = F^{\text{fnt}}(c_{\text{tex}}, i_{\text{fnt}}).
\label{eq:preliminary_font}
\end{equation}
\noindent\textbf{Text removal and segmentation.}
Using the spatial feature $c_{\text{spa}}$, a removal head $F^{\text{rmv}}$ reconstructs a background image $\tilde{i}_{\text{bg}}$ by removing the text, and a segmentation head $F^{\text{seg}}$ produces the binary text region mask $\tilde{i}_{\text{seg}}$.
\begin{equation}
\tilde{i}_{\text{bg}} = F^{\text{rmv}}(c_{\text{spa}}), \tilde{i}_{\text{seg}} = F^{\text{seg}}(c_{\text{spa}}).
\label{eq:preliminary_bg_seg}
\end{equation}

The total loss is defined as a weighted sum of task-specific losses:
\begin{equation}
\begin{split}
L =\ & L_{\text{clr}}(\tilde{i}_{\text{clr}}, i^{\text{gt}}_{\text{clr}}) + L_{\text{fnt}}(\tilde{i}_{\text{fnt}}, i^{\text{gt}}_{\text{fnt}}) \\
& + L_{\text{bg}}(\tilde{i}_{\text{bg}}, i^{\text{gt}}_{\text{bg}}) + L_{\text{seg}}(\tilde{i}_{\text{seg}}, i^{\text{gt}}_{\text{seg}})
\label{eq:preliminary_loss}
\end{split}
\end{equation}

Each loss is computed with appropriate metrics (e.g., MSE for $L_{\text{clr}}$, MAE for $L_{\text{bg}}$, Dice loss~\cite{diceloss} for $L_{\text{fnt}}$, $L_{\text{seg}}$), supervised by synthetic ground truths. This task-oriented pre-training enables text style encoder to extract interpretable and transferable style features.

The extracted glyph features $C_{\text{struct}}$ and style features $C_{\text{style}}$ are used to guide the diffusion generator $G$ during training. Specifically, $C_{\text{glyph}}$ is used as key-value in the cross-attention layers for accurate text rendering, while the style features $C_{\text{style}}$ are applied to the skip connections and middle blocks to support high-fidelity rendering.

\subsection{Overview of STELLAR Framework}

We propose STELLAR, a framework designed to enhance STE performance for low-resource languages and real-world images through a multi-stage training strategy. Specifically, as illustrated in Figure~\ref{fig:main}, STELLAR guides a diffusion generator $G$ using glyph features $C_\text{glyph}$ and style features $C_\text{style}$.

To support multilingual adaptability, we introduce a language-adaptive glyph encoder $T$ that is trained with language-specific recognizers. By leveraging supervision from OCR models specialized to each language, $T$ learns to generate glyph representations that reflect language-dependent structural features, enabling accurate text rendering even in low-resource languages. The text style encoder $S$, identical to the baseline, extracts style features $C_\text{style}$ to provide fine-grained style guidance.

We adopt a multi-stage training strategy to enhance domain adaptability. Stage 1 utilizes synthetic text image pairs to pre-train the model, establishing fundamental text rendering capabilities. Stage 2 employs fine-tuning with real-world text image pairs to enhance domain adaptability. This approach enables high-quality text editing without requiring additional post-hoc techniques.

\begin{table*}[t]
    \begin{adjustbox}{width=\textwidth, center}
    \setlength{\tabcolsep}{4pt}
    \centering
    \small
    \begin{tabular}{l|cccc|cccc|cccc}
    \toprule
    & \multicolumn{4}{c|}{Korean} & \multicolumn{4}{c|}{Arabic} & \multicolumn{4}{c}{Japanese} \\
    \cmidrule(lr){2-5} \cmidrule(lr){6-9} \cmidrule(lr){10-13}
    Metric & AnyText & AnyText2 & TextFlux & \textbf{STELLAR}
           & AnyText & AnyText2 & TextFlux & \textbf{STELLAR}
           & AnyText & AnyText2 & TextFlux & \textbf{STELLAR} \\
    \midrule
    SSIM ($\uparrow$) & 0.2822 & 0.2626 & 0.3409 & \textbf{0.5061}
            & 0.3693 & 0.3230 & 0.3745 & \textbf{0.4311}
            & 0.2755 & 0.2696 & \textbf{0.4105} & 0.3520 \\
    PSNR ($\uparrow$) & 13.2072 & 12.6499 & 14.5132 & \textbf{16.1514}
            & 12.8449 & 11.5576 & 13.1960 & \textbf{13.9638}
            & 12.0874 & 10.0943 & \textbf{15.1957} & 13.9301 \\
    MSE ($\downarrow$) & 0.0577 & 0.0881 & 0.0457 & \textbf{0.0301}
            & 0.0631 & 0.1076 & 0.0589 & \textbf{0.0485}
            & 0.0760 & 0.1567 & \textbf{0.0427} & 0.0517 \\
    FID ($\downarrow$) & 122.4377 & 96.4728 & \textbf{24.9410} & 34.4719
            & 119.7389 & 117.4138 & \textbf{52.8297} & 62.8025
            & 148.7926 & 167.1994 & \textbf{69.4484} & 73.7080 \\
    \midrule
    $s_{\text{clr}}$($\uparrow$) & 0.7801 & 0.8337 & 0.9159 & \textbf{0.9188}
            & 0.7645 & 0.7514 & 0.8893 & \textbf{0.9254}
            & 0.7703 & 0.7235 & \textbf{0.8911} & 0.8857 \\
    $s_{\text{fnt}}$($\uparrow$) & 0.6952 & 0.6979 & 0.8229 & \textbf{0.8274}
            & 0.7129 & 0.6961 & 0.7908 & \textbf{0.8745}
            & 0.7000 & 0.7079 & 0.8052 & \textbf{0.8259} \\
    $s_{\text{bg}}$($\uparrow$) & 0.6928 & 0.6203 & 0.8004 & \textbf{0.8325}
            & 0.5951 & 0.5538 & 0.7346 & \textbf{0.7803}
            & 0.4475 & 0.3184 & \textbf{0.6608} & 0.6025 \\
    TAS ($\uparrow$) & 0.7227 & 0.7173 & 0.8464 & \textbf{0.8596}
            & 0.6908 & 0.6671 & 0.8049 & \textbf{0.8601}
            & 0.6393 & 0.5833 & \textbf{0.7857} & 0.7714 \\
    \midrule
    Rec.Acc ($\uparrow$) & 0.0010 & 0.0899 & 0.2213 & \textbf{0.8042}
            & 0.0000 & 0.0082 & 0.0714 & \textbf{0.6840}
            & 0.0000 & 0.1013 & 0.4156 & \textbf{0.4338} \\
    NED ($\uparrow$) & 0.0116 & 0.2796 & 0.4836 & \textbf{0.9115}
            & 0.0054 & 0.0576 & 0.4449 & \textbf{0.8985}
            & 0.0033 & 0.1989 & 0.6331 & \textbf{0.6356} \\
    
    \bottomrule
    \end{tabular}
    \end{adjustbox}
    \caption{Quantitative evaluation of STELLAR and baselines on the STIPLAR evaluation set.}
    \label{tab:main_metrics}
\end{table*}

\subsection{Language-Adaptive Glyph Encoding}

To accurately encode glyph information for diverse languages, STELLAR addresses the limitations of the text glyph structure encoder in baseline methods. The encoder learns glyph structures by aligning character-level features with visual features extracted by a scene text recognizer~\cite{abinet}, supervised via a CLIP~\cite{clip} loss. However, its capacity is limited because the recognizer is primarily pre-trained on English. To overcome this, we design a language-adaptive glyph encoder $T$, shown in Figure~\ref{fig:main}, that independently generates glyph features specialized for each language, using language-specific pre-trained recognizers~\cite{ppocrv4}. This modular design scales to new scripts and effectively captures complex visual properties such as character composition and spatial arrangement. For example, it accommodates the right-to-left writing direction and context-dependent letter forms in Arabic. Consequently, STELLAR performs accurate editing with high fidelity in multilingual contexts.

\subsection{Multi-Stage Training Strategy}

As shown in Figure~\ref{fig:main}, we adopt a multi‑stage training strategy that leverages both synthetic and real‑world data to secure robust domain generalization. This design enables STELLAR to balance text recognition accuracy, visual consistency, and generation quality even on real‑world images.

In Stage 1, diffusion generator $G$ is pre-trained on large-scale synthetic data, enabling it to learn basic rendering abilities while preserving structural integrity and style consistency. We filter high-quality image pairs using pre-trained recognizers~\cite{ppocrv4} to ensure precise text recognition, experimentally validating the benefits of using clear data to enhance recognition accuracy and robustness (see Appendix). We independently organize and train data for each language to capture language‑specific font styles and writing systems, improving adaptability across Korean, Arabic, and Japanese.

In Stage 2, we fine‑tune the pre‑trained generator $G$ on real‑world multilingual text image pairs from Korean, Arabic, and Japanese data that capture realistic visual complexities such as noise, texture, and lighting variation. Despite using less than 5\% of the synthetic data and 10\% of the training epochs, this stage rapidly adapts the model to real‑world domains, achieving improved editing performance.

\section{Dataset}

To improve adaptation to real-world scenarios, we constructed \textbf{STIPLAR} (\textbf{S}cene \textbf{T}ext \textbf{I}mage \textbf{P}airs of \textbf{L}ow-resource l\textbf{\text{A}}nguages and \textbf{R}eal-world data), a dataset of real text image pairs in Korean, Arabic, and Japanese. Existing datasets mostly focus on English or Chinese and rarely provide style-consistent pairs for low‑resource languages. We therefore collect and refine text images from two sources: open-source datasets and web crawling, and split it into training and evaluation sets (8:2 ratio).

From the MLT-2019~\cite{icdarmlt2019} training set of text crops, we select 1,000 images per language, discard low-quality samples, and have annotators correct label mismatches and manually annotate text pairs that share the same style and background but differ in text. This process yields 1,818 Korean, 2,328 Arabic, and 453 Japanese pairs.

In addition, we search Creative Commons–licensed images using queries in each language and English generated with GPT‑4o~\cite{gpt4o}. After OCR detection, cropping, quality filtering, safety checking, and privacy masking, we construct 7,946 Korean, 3,988 Arabic, and 1,570 Japanese pairs.

In total, STIPLAR contains 9,764 Korean, 6,316 Arabic, and 2,023 Japanese pairs across various domains such as signs, posters, menus, book covers, and other natural scenes. Designed for fine-tuning on real-world images, the dataset addresses challenges like background complexity, lighting variation, and distortion, offering a new benchmark for evaluating STE in low-resource languages. Further construction details and sample examples are in the Appendix.

\section{Text Appearance Similarity}

Most existing STE methods employ image similarity metrics such as SSIM, PSNR, MSE, and FID to evaluate quality of edited images. While these metrics provide an indirect proxy for style preservation, they suffer from two critical limitations. First, they assume the text content is unchanged, making evaluation infeasible in the absence of ground truth images, which is a common scenario for real-world data. Second, similarity scores alone offer little insight into whether differences arise from font, color, or background changes.

To overcome these limitations, we propose \textbf{T}ext \textbf{\text{A}}ppearance \textbf{S}imilarity (\textbf{TAS}), a novel metric specifically designed to intuitively and accurately assess visual style similarity in text images. TAS separately assesses the similarity of visual attributes such as color, font, and background, and then averages these individual measurements to produce a comprehensive score. As a result, TAS can evaluate edited images without ground truth references, providing greater interpretability beyond mere similarity scores.

Given two images $I_\text{A}$ and $I_\text{B}$ with text strings $t_\text{A}$ and $t_\text{B}$, we extract the texture features $c_{\text{tex}}$ and spatial features $c_{\text{spa}}$ from text style encoder $S$ as defined in Equation~\ref{eq:preliminary_tex_spa}. For each image, we obtain $(c^{\text{A}}_{\text{tex}}, c^{\text{A}}_{\text{spa}})$ and $(c^{\text{B}}_{\text{tex}}, c^{\text{B}}_{\text{spa}})$.

\noindent\textbf{Color similarity $s_{\text{clr}}$.}
As described in Equation~\ref{eq:preliminary_color}, the texture features $c^{\text{A}}_{\text{tex}}$ and $c^{\text{B}}_{\text{tex}}$ are applied to a grayscale image $i_{\text{clr}}$ rendered from $t_\text{B}$, resulting in colorized images $\tilde{i}^{\text{A}}_{\text{clr}}$ and $\tilde{i}^{\text{B}}_{\text{clr}}$. The similarity is measured using the normalized CIEDE2000~\cite{ciede2000} metric:
\begin{equation}
s_{\text{clr}} = 1 - \min\left(\frac{\operatorname{CIEDE2000}(\tilde{i}^{\text{A}}_{\text{clr}}, \tilde{i}^{\text{B}}_{\text{clr}})}{50}, 1\right).
\label{eq:tas_color}
\end{equation}

\noindent\textbf{Font similarity $s_{\text{fnt}}$.}
As described in Equation~\ref{eq:preliminary_font}, the texture features $c^{\text{A}}_{\text{tex}}$ and $c^{\text{B}}_{\text{tex}}$ are applied to a template font glyph image $i_{\text{fnt}}$ to generate the reshaped images $\tilde{i}^{\text{A}}_{\text{fnt}}$ and $\tilde{i}^{\text{B}}_{\text{fnt}}$. Their similarity is computed using FSIM~\cite{fsim}:
\begin{equation}
s_{\text{fnt}} = \operatorname{FSIM}(\tilde{i}^{\text{A}}_{\text{fnt}}, \tilde{i}^{\text{B}}_{\text{fnt}}).
\label{eq:tas_font}
\end{equation}

\noindent\textbf{Background similarity $s_{\text{bg}}$.}
As described in Equation~\ref{eq:preliminary_bg_seg}, the spatial features $c_{\text{spa}}$ are used to reconstruct the background images $\tilde{i}^{\text{A}}_{\text{bg}}$ and $\tilde{i}^{\text{B}}_{\text{bg}}$ from $I_\text{A}$ and $I_\text{B}$, respectively. We compute similarity via MS-SSIM~\cite{msssim}:
\begin{equation}
s_{\text{bg}} = \operatorname{MS\text{-}SSIM}(\tilde{i}^{\text{A}}_{\text{bg}}, \tilde{i}^{\text{B}}_{\text{bg}}).
\label{eq:tas_bg}
\end{equation}

\begin{figure}[t]
\centering
\includegraphics[width=\columnwidth]{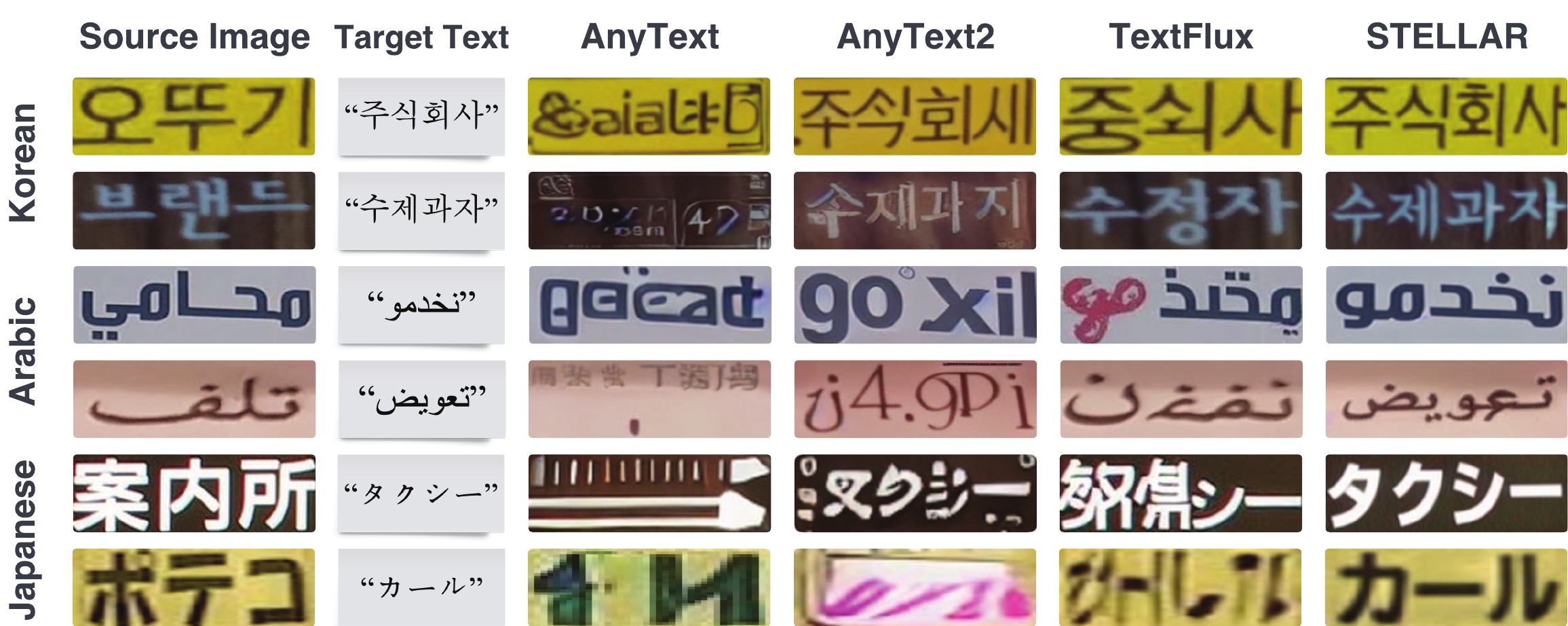}
\caption{Comparison of edited results across baselines on Korean, Arabic, and Japanese text images.}
\label{fig:qualitative}
\end{figure}

\begin{table}[t]
\begin{adjustbox}{width=\columnwidth, center}
\setlength{\tabcolsep}{8pt}
\centering
\small
\begin{tabular}{l|ccccc}
\toprule
\textbf{Metric} 
& \makecell{\ding{51} T \\ \ding{55} F} 
& \makecell{\ding{51} T \\ \ding{55} C} 
& \makecell{\ding{51} T \\ \ding{55} B} 
& \makecell{\ding{51} T \\ \ding{55} F/C/B} 
& \makecell{\ding{55} T \\ \ding{51} F/C/B} \\

\midrule
SSIM ($\uparrow$) & 0.6379 & \textbf{0.7783} & 0.5653 & 0.3859 & 0.5768 \\
PSNR ($\uparrow$) & 18.0882 & \textbf{21.0446} & 10.2451 & 8.6073 & 17.2873 \\
MSE ($\downarrow$) & 0.0263 & \textbf{0.0230} & 0.1513 & 0.1791 & 0.0306 \\
FID ($\downarrow$) & 24.9213 & \textbf{21.2495} & 41.7723 & 42.6132 & 27.2827 \\
\midrule
$s_{\text{clr}}$ ($\uparrow$) & 0.8974 & 0.6477 & 0.8744 & 0.6339 & \textbf{0.9341} \\
$s_{\text{fnt}}$ ($\uparrow$) & 0.6766 & \textbf{0.9044} & 0.8759 & 0.6706 & 0.8172 \\
$s_{\text{bg}}$ ($\uparrow$) & 0.9381 & \textbf{0.9614} & 0.3726 & 0.3619 & 0.9268 \\
TAS ($\uparrow$) & 0.8374 & 0.8379 & 0.7076 & 0.5555 & \textbf{0.8933} \\
\bottomrule
\end{tabular}
\end{adjustbox}
\caption{Quantitative results on synthetically manipulated Korean image pairs with controlled variations in text or style attributes. Each column corresponds to a specific variation: T(text content), F(font), C(color), B(background). \ding{51} and \ding{55} indicate preserved and modified attributes, respectively.}
\label{tab:variation_metrics}
\end{table}

The final TAS score averages these three similarities:
\begin{equation}
\operatorname{TAS}\!\bigl(I_\text{A}, I_\text{B}\!\bigr) = \frac{(s_{\text{clr}} + s_{\text{fnt}} + s_{\text{bg}})}{3}.
\label{eq:tas}
\end{equation}

\begin{table}[t]
\begin{adjustbox}{width=\columnwidth, center}
\setlength{\tabcolsep}{4pt}
\centering
\small
\begin{tabular}{l|cc|cc|cc}
\toprule
& \multicolumn{2}{c|}{Korean} 
& \multicolumn{2}{c|}{Arabic} 
& \multicolumn{2}{c}{Japanese} \\
\cmidrule(lr){2-3} \cmidrule(lr){4-5} \cmidrule(lr){6-7}
Metric & w/ source & w/ GT 
& w/ source & w/ GT 
& w/ source & w/ GT \\
\midrule
SSIM($\uparrow$) & 0.3063 & \textbf{0.5295}
                & 0.4169 & \textbf{0.4391}
                & 0.3872 & \textbf{0.4007} \\
PSNR($\uparrow$) & 13.6226 & \textbf{16.3848}
                & 13.4986 & \textbf{14.0340}
                & 13.8947 & \textbf{14.2847} \\
MSE($\downarrow$) & 0.0521 & \textbf{0.0282}
                 & 0.0505 & \textbf{0.0479}
                 & 0.0514 & \textbf{0.0475} \\
FID($\downarrow$) & 47.3129 & \textbf{45.9684}
                 & 84.9011 & \textbf{82.2361}
                 & 111.0451 & \textbf{110.5366} \\
\midrule
TAS($\uparrow$) & 0.8608 & \textbf{0.8641}
               & \textbf{0.8726} & 0.8636
               & \textbf{0.8031} & 0.7627 \\
\bottomrule
\end{tabular}
\end{adjustbox}
\caption{Quantitative results of STELLAR evaluated against source (w/ source) and ground truth (w/ GT) images.}
\label{tab:gt_vs_src}
\end{table}

\section{Experiments}

\subsection{Implementation Details}

\noindent\textbf{Baselines.}
We compare STELLAR exclusively with three mask‑and‑inpaint methods that support low-resource or multilingual text editing, as other STE models fail to generate valid outputs in these languages. All three baselines receive the full image and a mask as input, and AnyText~\cite{anytext} and AnyText2~\cite{anytext2} additionally accept a text prompt. For a fair comparison, we crop a fixed‑size patch from each image, create the corresponding mask, and supply each model with its required inputs. After generation, we crop the edited text region from every output for evaluation.

\noindent\textbf{Training and Evaluation.}
Stage 1 pre‑trains the diffusion generator on 200k synthetic text image pairs~\cite{synthtext} per language (Korean, Japanese, Arabic), using only OCR-correct pairs. Stage 2 fine‑tunes the model on the STIPLAR dataset. We use the pre-trained checkpoint of Stable Diffusion~\cite{ldm} v1.5, resize all source images to 256$\times$256, and set the maximum target text length to 12. Training is conducted with a learning rate of $1\times10^{-5}$ using 2 NVIDIA H100 80GB GPUs, training Stage 1 for 100 epochs (66 hours) and Stage 2 for 10 epochs (0.3 hours). 

We evaluate image quality using SSIM, PSNR, MSE, and FID, style preservation with TAS, and text correctness with OCR recognition accuracy(Rec.Acc) and Normalized Edit Distance(NED).

\begin{table*}[t]
\begin{adjustbox}{width=\textwidth, center}
\setlength{\tabcolsep}{6pt}
\centering
\small
\begin{tabular}{l|cccc|cccc|ccc}
\toprule
\textbf{Metric} 
& \multicolumn{4}{c|}{Korean} 
& \multicolumn{4}{c|}{Arabic} 
& \multicolumn{3}{c}{Japanese} \\
\cmidrule(lr){2-5} \cmidrule(lr){6-9} \cmidrule(lr){10-12}
& S1 & S1 w/PH & S1 + S2 w/DS & \textbf{STELLAR} 
& S1 & S1 w/PH & S1 + S2 w/DS & \textbf{STELLAR} 
& S1 & S1 w/PH & \textbf{STELLAR} \\
\midrule
SSIM($\uparrow$) & 0.3788 & 0.3808 & 0.5000 & \textbf{0.5061} & 0.3668 & 0.3747 & 0.4277 & \textbf{0.4311} & 0.3121 & 0.3140 & \textbf{0.3520} \\
PSNR($\uparrow$) & 14.7164 & 14.7939 & 15.9362 & \textbf{16.1514} & 13.2688 & 13.4965 & 13.8625 & \textbf{13.9638} & 13.5722 & 13.6210 & \textbf{13.9301} \\
MSE($\downarrow$) & 0.0427 & 0.0420 & 0.0317 & \textbf{0.0301} & 0.0562 & 0.0537 & 0.0493 & \textbf{0.0485} & 0.0567 & 0.0563 & \textbf{0.0517} \\
FID($\downarrow$) & 36.1512 & 35.8567 & 36.9030 & \textbf{34.4719} & 62.7250 & \textbf{61.6446} & 62.2387 & 62.8025 & 76.4246 & 76.1548 & \textbf{73.7080} \\
\midrule
$s_{\text{clr}}$($\uparrow$) & \textbf{0.9198} & 0.9196 & 0.9181 & 0.9188 & 0.9241 & \textbf{0.9266} & 0.9199 & 0.9254 & \textbf{0.8877} & 0.8867 & 0.8857 \\
$s_{\text{fnt}}$($\uparrow$) & 0.8141 & 0.8147 & 0.8215 & \textbf{0.8274} & 0.8319 & 0.8335 & 0.8730 & \textbf{0.8745} & 0.8303 & \textbf{0.8308} & 0.8259 \\
$s_{\text{bg}}$($\uparrow$) & 0.7748 & 0.7781 & 0.8233 & \textbf{0.8325} & 0.7363 & 0.7399 & 0.7756 & \textbf{0.7803} & 0.5792 & 0.5805 & \textbf{0.6025} \\
TAS($\uparrow$) & 0.8362 & 0.8375 & 0.8543 & \textbf{0.8596} & 0.8308 & 0.8333 & 0.8562 & \textbf{0.8601} & 0.7657 & 0.7660 & \textbf{0.7714} \\
\midrule
Rec. ACC($\uparrow$) & 0.6676 & 0.6710 & 0.7710 & \textbf{0.8042} & 0.6412 & 0.6290 & 0.6799 & \textbf{0.6840} & 0.2987 & 0.2961 & \textbf{0.4338} \\
NED($\uparrow$) & 0.8443 & 0.8452 & 0.8974 & \textbf{0.9115} & 0.8375 & 0.8300 & 0.8912 & \textbf{0.8985} & 0.5197 & 0.5169 & \textbf{0.6356} \\
\bottomrule
\end{tabular}
\end{adjustbox}
\caption{Quantitative comparison of four configurations of STELLAR: S1 (Stage 1; pre-training only), S1 w/ PH (post-hoc technique), S1 + S2 w/ DS (Stage 2 fine-tuning with downsampled real data, applied to Korean and Arabic only), and full model fine-tuned on full real-world data.}
\label{tab:ablation_metrics}
\end{table*}

\subsection{Quantitative Results}

Table~\ref{tab:main_metrics} compares STELLAR with three multilingual baselines on the STIPLAR evaluation set. STELLAR achieves the best overall performance in Korean and Arabic, and remains competitive in Japanese.

For image quality, STELLAR consistently outperforms baselines across most metrics for Korean and Arabic. It achieves the highest TAS scores in both languages, reflecting superior preservation of text style and background. In Japanese, however, TextFlux~\cite{textflux} slightly outperforms STELLAR in terms of image quality and style preservation, likely due to the visual similarity between Japanese kanji and the Chinese characters prevalent in the training data of the baselines. Despite this, STELLAR achieves the highest recognition accuracy across all languages. For Korean, it records 0.8042 in Rec.Acc and 0.9115 in NED, outperforming TextFlux by absolute margin of 0.5829 and 0.4279, respectively. Even in Japanese, STELLAR maintains its advantage in recognition accuracy.

To further validate robustness, we additionally evaluated STELLAR on external real-world benchmarks without ground truth references (see Appendix). The model maintained consistently strong performance, confirming its ability to generalize beyond the STIPLAR evaluation set.

These results show that STELLAR offers well‑balanced improvements in image quality, style preservation, and recognition accuracy for multiple low‑resource languages in real-world scenarios.

\subsection{Qualitative Results}

Qualitative comparisons using text images from the STIPLAR evaluation set revealed that STELLAR produced visually consistent and clear edited results, as shown in Figure~\ref{fig:qualitative}. AnyText and AnyText2 frequently generated incomplete or unclear text, resulting in noticeable quality degradation. TextFlux showed relatively stable visual style and background consistency, attributable to the strong contextual reasoning of its DiT‑based generative model~\cite{flux1}, but often struggled with textual clarity and alignment. Conversely, STELLAR consistently produced edited text images that closely matched the source image’s style and background while rendering characters accurately and without noticeable distortions. Additional qualitative examples are provided in the Appendix.

\subsection{TAS Analysis}

\noindent\textbf{Validating TAS on Visual Variations.}
We created five synthetic Korean datasets, with controlled variation in text content, font, color, or background. As shown in Table~\ref{tab:variation_metrics}, when only the text content differed, SSIM and PSNR recorded relatively low scores (0.5768 and 17.2873, respectively), while TAS achieved the highest score (0.8933), demonstrating its robustness to text changes.

Conversely, in the color-modified set, conventional metrics such as SSIM (0.7783), PSNR (21.0446), MSE (0.0230), and FID (21.2495) reported high similarity scores, whereas TAS yielded a comparatively lower score (0.8379). This indicates TAS's heightened sensitivity to style changes and confirms its reliability as a style-centric metric. Moreover, because TAS separately measures color, font, and background similarity, it provides interpretable feedback by revealing which aspect of style has changed. We assess the reliability of TAS through its high correlation with human evaluations, and further validate its effectiveness on other languages. The experiments are detailed in the Appendix.

\noindent\textbf{Evaluation without Ground Truth.}
To determine whether TAS can evaluate visual style preservation without ground truth images, we filtered samples where the text was accurately rendered using OCR and then conducted evaluation between the source and generated images. As shown in Table~\ref{tab:gt_vs_src}, TAS yielded comparable or even higher scores than ground truth-based comparisons in all languages, despite lower SSIM, PSNR, MSE, and FID scores. These findings indicate that TAS reliably measures visual style preservation despite changes in text content and is therefore practical for scenarios where reference images cannot be provided.

\subsection{Ablation Study}

\noindent\textbf{Analysis of Multi-Stage Training.}
Stage 2 (S2) fine-tuning plays a critical role in real-world image domain adaptation. As shown in Table~\ref{tab:ablation_metrics}, models trained solely on Stage 1 (S1) exhibited degraded performance in image quality, TAS, and recognition accuracy across all languages. Even with only S1 training, the model maintains higher recognition accuracy in Korean and Arabic compared to baselines reported in Table~\ref{tab:main_metrics}. Considering that all baselines are primarily trained on Chinese characters, this suggests that our language-specific training strategy contributes to improved text recognition accuracy. Examples of generated results for each case can be found in the Appendix.

\noindent\textbf{Comparison with Post-hoc Techniques.}
Previous research~\cite{textctrl} tackles the domain gap that occurs when models trained on synthetic data are deployed on real-world images by applying a post‑hoc (PH) technique that injects source image features during inference. As shown in Table~\ref{tab:ablation_metrics}, applying this technique to a S1‑only model still yielded lower scores than STELLAR across all languages. This demonstrates that fine-tuning with real-world image pairs is more effective for domain adaptation than post-hoc adjustments.

\noindent\textbf{Analysis of Dataset Size in Stage 2.}
To investigate the impact of dataset size on S2 fine-tuning, we fine-tuned the Korean and Arabic models in S2 datasets randomly downsampled (DS) to match the size of the Japanese dataset. The results in Table~\ref{tab:ablation_metrics} show that S1 + S2 w/DS still outperforms TextFlux reported in Table~\ref{tab:main_metrics}, but performs worse than STELLAR in most metrics. This indicates that a larger amount of real-world data in S2 leads to better performance, while the consistent improvements over the S1-only model confirm the effectiveness of the multi-stage training strategy.

\section{Conclusion and Limitations}

With the growing demand for multilingual STE in practical applications such as AR signage translation and game or film localization, robust STE solutions are increasingly important. We propose STELLAR, a robust framework that supports low-resource languages and real-world adaptation through a language-adaptive glyph encoder and multi-stage training. STELLAR demonstrates strong performance in terms of image quality, visual style preservation, and recognition accuracy. We further introduce STIPLAR, a real-world text image pair dataset for low-resource languages, and TAS, a new metric for evaluating style preservation, both of which validate the practicality and effectiveness of our approach.

However, our study has several limitations. The limited size of real-world datasets and restricted language coverage may hinder generalization, especially under various noise or uncommon styles. Editing performance also declines for longer text inputs due to the scarcity of long-text samples in the collected datasets, and related failure cases are analyzed in the Appendix. We plan to expand language diversity and collect more diverse data, and exploring unsupervised domain adaptation and zero-shot editing for unseen scripts to improve scalability and real-world applicability.

\section*{Acknowledgments}

This work was supported by the IITP (Institute of Information \& Communications Technology Planning \& Evaluation)-ICAN (ICT Challenge and Advanced Network of HRD) (IITP-2024-RS-2023-00259806, 20\%) grant funded by the Korea government(Ministry of Science and ICT) and the National Research Foundation of Korea (NRF) grant funded by the Korea government (MSIT) (RS-2024-00354675, 40\%) and (RS-2024-00352184, 40\%).

\bibliography{aaai2026}

\clearpage

\appendix
\section*{Appendix}

\setcounter{secnumdepth}{2}
\renewcommand{\thetable}{A\arabic{table}}
\renewcommand{\thefigure}{A\arabic{figure}}
\setcounter{table}{0}
\setcounter{figure}{0}

\section{Implementation Details}

\subsection{Architectural Components}

The core component of the STELLAR framework is the language-adaptive glyph encoder $T$, implemented using a lightweight transformer~\cite{transformer}. For training $T$, we leverage language-specific recognizers from the pre-trained PPOCRv4~\cite{ppocrv4} (\texttt{korean\allowbreak\_PP-OCRv4\allowbreak\_rec\allowbreak\_infer}, \texttt{arabic\allowbreak\_PP-OCRv4\allowbreak\_rec\allowbreak\_infer}, and \texttt{japan\allowbreak\_PP-OCRv4\allowbreak\_rec\allowbreak\_infer}).

Text style encoder $S$ is built on a ViT-B~\cite{vit} backbone and trained to disentangle four visual attributes including text color, font, background, and spatial mask, through four dedicated heads. The modules for reconstructing text styles are as follows:

\begin{itemize}
    \item $F^{\text{clr}}$: ResNet34~\cite{resnet} backbone with Adaptive Instance Normalization~\cite{adain} and a multi-scale decoder with upsampling layers.
    \item $F^{\text{fnt}}$: ResNet34 backbone with Pyramid Pooling Module~\cite{ppm} and multi-scale decoder with upsampling layers.
    \item $F^{\text{rmv}}$, $F^{\text{seg}}$: Residual convolution blocks integrated with spatial attention~\cite{cbam}.
\end{itemize}

\subsection{Training Configuration}

All training was conducted on \texttt{Ubuntu 24.04} with Python 3.10.12, using 2 NVIDIA H100 80GB GPUs. The software environment is detailed in Table~\ref{tab:library_versions}.

The hyperparameters for each module are summarized in Table~\ref{tab:hyperparameters}, and we use the AdamW~\cite{adamw} optimizer with $\beta{=}(0.9, 0.999)$ and a weight decay of 0.01 throughout training.

\subsection{Inference and Evaluation Protocol}

Inference and evaluation are conducted under the same hardware setup. Sampling parameters include 50 denoising steps, classifier-free guidance scale 2.0, and random seed 42. 

We use official GitHub checkpoints of the baseline models (AnyText~\cite{anytext}, AnyText2~\cite{anytext2}, TextFlux~\cite{textflux}). During inference, each image is cropped into a fixed-size patch (up to 1024 $\times$ 1024) so that the target text region is positioned near the center, and the corresponding mask is created before being fed into the models. We adopt this cropping strategy because in many cases, the original images are considerably larger, while the target text occupies only a small portion of the image. Such imbalance often leads to degraded editing performance. Therefore, cropping a fixed-size patch helps maintain consistent input scales and ensures stable editing quality. After generation, the edited text region is cropped from each output for evaluation.

\begin{itemize}
    \item Text prompts for AnyText and AnyText2 are generated using the GPT-4o~\cite{gpt4o} (version \texttt{gpt-4o-2024-08-06}).
    \item OCR evaluation: PPOCRv4 is used for Korean and Japanese, Google Cloud Vision API~\cite{googlecloudvision} is used for Arabic.
\end{itemize}

\subsection{Post-hoc Method Comparison}

We adapt the post-hoc technique proposed in TextCtrl~\cite{textctrl}, specifically GaMuSa (Glyph-adaptive Mutual Self-Attention), to the three target languages by replacing its original vision encoder with the corresponding PPOCRv4 recognizer. This enables fair comparison on Korean, Japanese, and Arabic datasets.

\begin{table}[t]
\setlength{\tabcolsep}{4pt}
\centering
\small
\begin{tabular}{l|c}
\toprule
Library & Version \\
\midrule
torch & 2.1.0a0+32f93b1 \\
torchvision & 0.16.0a0 \\
pytorch-lightning & 1.9.1 \\
accelerate & 0.30.0 \\
transformers & 4.41.0 \\
paddleocr & 2.10.0 \\
google-cloud-vision & 3.10.2 \\
\bottomrule
\end{tabular}
\caption{Library versions used in our implementation environment.}
\label{tab:library_versions}
\end{table}

\begin{table}[t]
\setlength{\tabcolsep}{4pt}
\centering
\small
\begin{tabular}{l|cccc}
\toprule
Module & Learning rate & Batch size & Epochs & Image size \\
\midrule
$T$ & 0.0001 & 256 & 100 & 48 $\times$ 320 \\
$S$ & 0.00001 & 256 & 80 & 128 $\times$ 128 \\
$G$ & 0.00001 & 128 & 100/10 & 256 $\times$ 256 \\
\bottomrule
\end{tabular}
\caption{Training hyperparameters for each module in STELLAR. Image sizes denote the resolution at which inputs are resized: for language-adaptive glyph encoder $T$, this refers to the input size of the language-specific recognizer, while for text style encoder $S$ and diffusion generator $G$, it corresponds to the resized input directly fed into the modules. Epochs for $G$ indicate those used in Stage 1 and Stage 2, respectively.}
\label{tab:hyperparameters}
\end{table}

\section{STIPLAR Dataset Construction}

\subsection{Data Collection and Annotation Pipeline}

\noindent\textbf{Open-Source Dataset Collection.}
As an initial step, we collect cropped text images from 1,000 full images for each language in the MLT-2019~\cite{icdarmlt2019} training set, yielding 9,497 Korean, 8,010 Arabic, and 10,460 Japanese samples. 

Only images containing exclusively the target language are retained. To ensure sufficient resolution, we discard samples with a pixel area (width $\times$ height) below 1,000 pixels and enforce a landscape orientation by keeping only images with width greater than height. For each full image, we generate text image pairs by enumerating all cropped text images within the same source. These pairs are then manually verified by language-proficient annotators, who perform \textbf{(1)} quality filtering, \textbf{(2)} correction of label mismatches, and \textbf{(3)} pairwise annotation to ensure consistent font, text color, and background, while differing in textual content.

\begin{table}[t]
\begin{adjustbox}{width=\columnwidth, center}
\centering
\small
\setlength{\tabcolsep}{2pt}
\begin{tabular}{ll|ccc|ccc}
\toprule
\multirow{2}{*}{\textbf{Lang.}} & \multirow{2}{*}{\textbf{Type}} 
& \multicolumn{3}{c|}{\textbf{Train}} & \multicolumn{3}{c}{\textbf{Eval}} \\
& & Open & Crawl & Total & Open & Crawl & Total \\
\midrule
\multirow{2}{*}{Korean} 
& Full image        & 269 & 317 & 586 & 68 & 80 & 148 \\
& Text image pair   & 1456 & 6229 & 7685 & 362 & 1717 & 2079 \\
\midrule
\multirow{2}{*}{Arabic} 
& Full image        & 251 & 55 & 306 & 64 & 15 & 79 \\
& Text image pair   & 1878 & 3457 & 5335 & 450 & 531 & 981 \\
\midrule
\multirow{2}{*}{Japanese} 
& Full image        & 97 & 252 & 349 & 25 & 63 & 88 \\
& Text image pair   & 356 & 1282 & 1638 & 97 & 288 & 385 \\
\bottomrule
\end{tabular}
\end{adjustbox}
\caption{STIPLAR dataset statistics by language(Lang.), image type, and data source. Open: open-source, Crawl: web-crawled.}
\label{tab:dataset_stats}
\end{table}

\noindent\textbf{Web-Crawled Dataset Collection.}
We collect additional images via Google Image Search using Creative Commons license filters. For each language (Korean, Arabic, Japanese), we generate 100 search queries, along with 100 English-based queries per language, using GPT-4o (version \texttt{gpt-4o-2024-08-06}). These queries span diverse categories including signage, menus, labels, instructions, book covers, and advertisements. After removing duplicates, we filter the images through the following stages:

\begin{itemize}
    \item Content filtering: GPT-4o is used to exclude non-natural scenes, as well as images containing profanity, explicit content, or political text.
    \item License filtering: We retain only images with licenses \texttt{CC0}, \texttt{CC BY}, \texttt{CC BY-NC}, or \texttt{PDM}.
    \item Privacy filtering: Annotators manually blur all faces and license plates to ensure privacy.
\end{itemize}

We then detect and crop text regions using the Upstage Document OCR API~\cite{upstageocr} for Korean and Japanese, and the Google Cloud Vision OCR API~\cite{googlecloudvision} for Arabic.

\begin{figure}[t]
\centering
\includegraphics[width=\columnwidth]{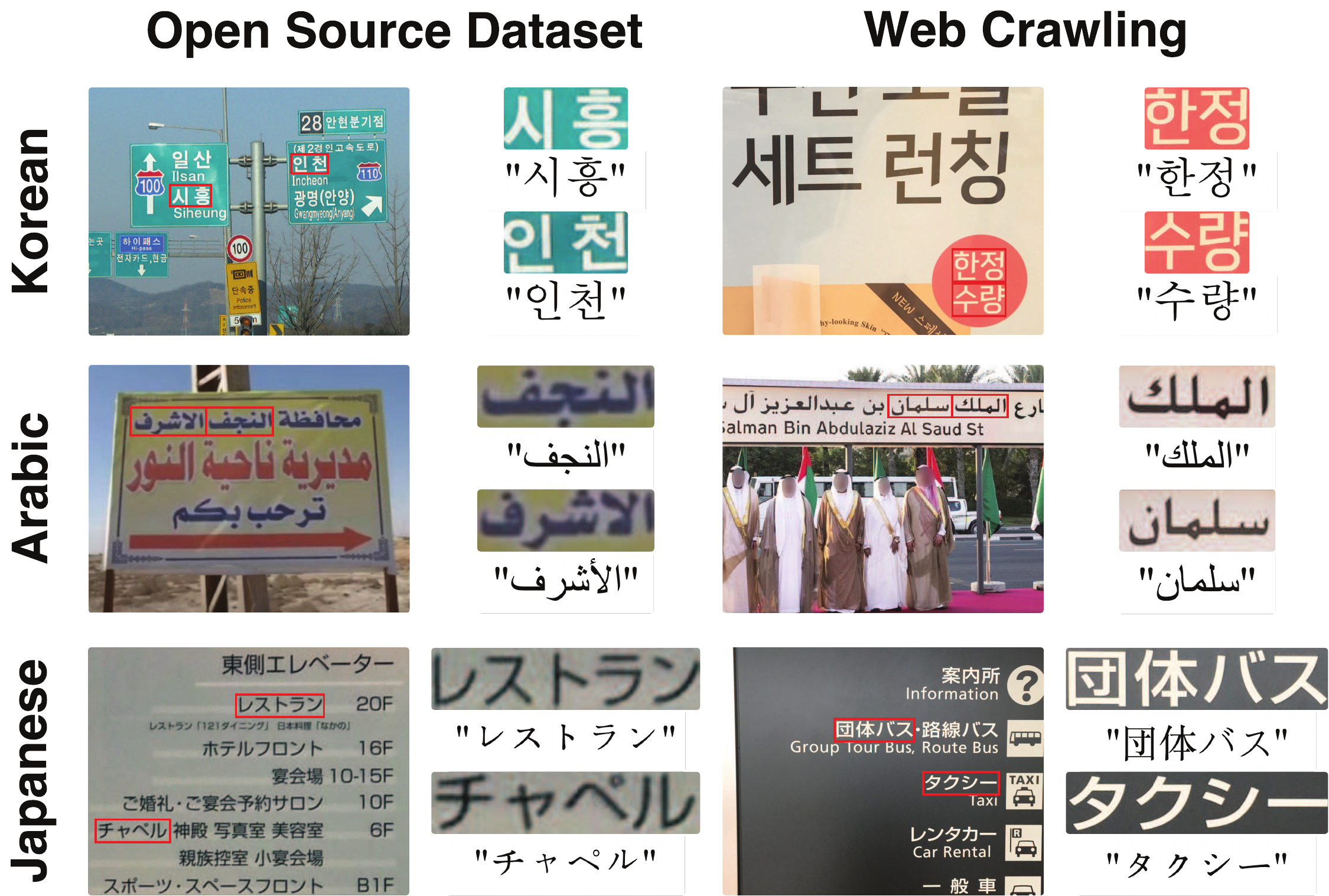}
\caption{Examples of full images and their corresponding cropped text image pairs in Korean, Arabic, and Japanese from the STIPLAR dataset.}
\label{fig:examples_stiplar}
\end{figure}

Only images meeting the same filtering criteria (minimum area $\geq$ 1000 pixels and landscape orientation) are preserved. As with the open-source data, image pairs are formed from text crops within each full image and annotated by language-proficient annotators following the same three-stage process: \textbf{(1)} quality filtering, \textbf{(2)} correction of label mismatches, and \textbf{(3)} pairwise annotation.

In total, we collect 18,107 image pairs across the three languages. To facilitate future research, we additionally release the original full images from which the text pairs were constructed. Table~\ref{tab:dataset_stats} summarizes dataset statistics. 

\subsection{Dataset Samples}

Figure~\ref{fig:examples_stiplar} presents representative examples of real-world text image pairs in Korean, Arabic, and Japanese from the STIPLAR dataset.

\begin{figure*}[t]
\centering
\includegraphics[width=0.75\textwidth]{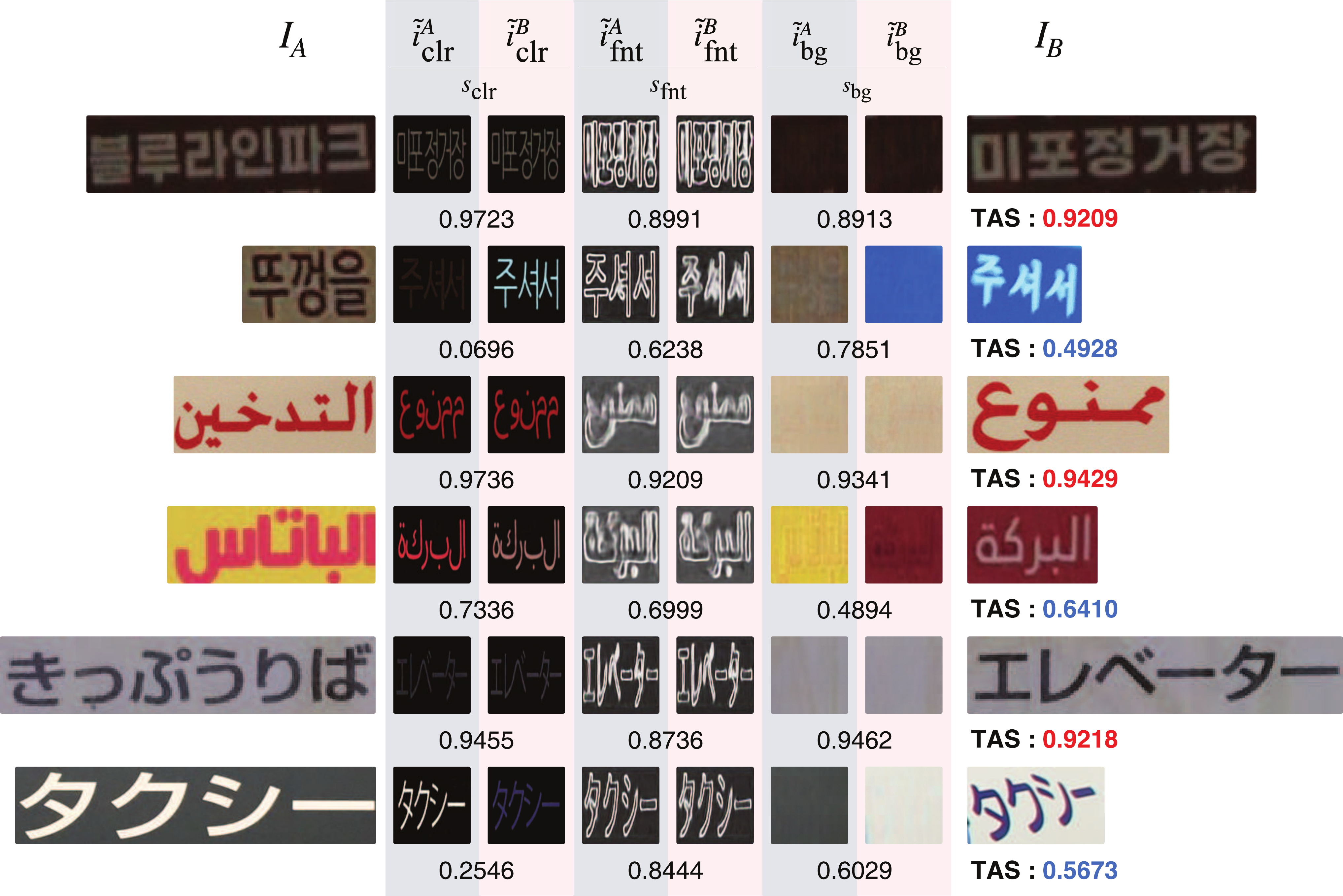}
\caption{
Examples used in TAS computation. Given an image pair $I_A$ and $I_B$, the text style encoder $S$ extracts colorized ($\tilde{i}^{\text{A}}_{\text{clr}}$, $\tilde{i}^{\text{B}}_{\text{clr}}$), font-reshaped ($\tilde{i}^{\text{A}}_{\text{fnt}}$, $\tilde{i}^{\text{B}}_{\text{fnt}}$), and text-removed ($\tilde{i}^{\text{A}}_{\text{bg}}$, $\tilde{i}^{\text{B}}_{\text{bg}}$) outputs. Each similarity score ($s_\text{clr}$, $s_\text{fnt}$, $s_\text{bg}$) is computed and averaged to obtain the final TAS score. In the third row (Arabic), although the text content differs, strong similarities in color, font, and background lead to a high TAS score of 0.9429. Conversely, in the sixth row (Japanese), even with identical text content, noticeable differences in color (0.2546) and background (0.6029) result in a lower TAS score of 0.5673, illustrating that TAS evaluates visual style consistency independently of textual content.}
\label{fig:tas_examples}
\end{figure*}

\begin{table}[t]
\begin{adjustbox}{width=\columnwidth, center}
\centering
\small
\setlength{\tabcolsep}{3pt}
\begin{tabular}{ll|ccccc}
\toprule
\textbf{Lang.} & \textbf{Metric} 
& \makecell{\ding{51}T\\\ding{55}F} 
& \makecell{\ding{51}T\\\ding{55}C} 
& \makecell{\ding{51}T\\\ding{55}B} 
& \makecell{\ding{51}T\\\ding{55}F/C/B} 
& \makecell{\ding{55}T\\\ding{51}F/C/B} \\
\midrule
\multirow{5}{*}{Arabic} 
& SSIM ($\uparrow$) & 0.6434 & \textbf{0.8006} & 0.5539 & 0.3913 & 0.6331 \\
& PSNR ($\uparrow$) & 17.6926 & \textbf{21.6005} & 9.7969 & 8.6104 & 17.9598 \\
& MSE ($\downarrow$) & 0.0260 & \textbf{0.0177} & 0.1560 & 0.1766 & 0.0257 \\
& FID ($\downarrow$) & 30.5648 & \textbf{23.2914} & 47.6726 & 47.4414 & 26.3165 \\
& TAS ($\uparrow$) & 0.8630 & 0.8361 & 0.7024 & 0.5649 & \textbf{0.9094} \\
\midrule
\multirow{5}{*}{Japanese} 
& SSIM ($\uparrow$) & 0.6843 & \textbf{0.7678} & 0.5070 & 0.3339 & 0.5551 \\
& PSNR ($\uparrow$) & 19.4541 & \textbf{20.6051} & 9.5679 & 8.4064 & 17.2124 \\
& MSE ($\downarrow$) & \textbf{0.0195} & 0.0219 & 0.1549 & 0.1791 & 0.0294 \\
& FID ($\downarrow$) & \textbf{24.6817} & 24.8031 & 46.9891 & 47.6065 & 31.9053 \\
& TAS ($\uparrow$) & 0.8858 & 0.8516 & 0.6917 & 0.5824 & \textbf{0.9071} \\
\bottomrule
\end{tabular}
\end{adjustbox}
\caption{Quantitative results from conventional metrics on synthetic image pairs with controlled variations in text (T), font (F), color (C), and background (B). Each column indicates whether the corresponding attributes are preserved (\ding{51}) or modified (\ding{55}).}
\label{tab:visual_variations_ar_jp}
\end{table}

\begin{table}[t]
\begin{adjustbox}{width=\columnwidth, center}
\centering
\small
\setlength{\tabcolsep}{3pt}
\begin{tabular}{ll|ccccc}
\toprule
\textbf{Lang.} & \textbf{Similarity} 
& \makecell{\ding{51}T\\\ding{55}F} 
& \makecell{\ding{51}T\\\ding{55}C} 
& \makecell{\ding{51}T\\\ding{55}B} 
& \makecell{\ding{51}T\\\ding{55}F/C/B} 
& \makecell{\ding{55}T\\\ding{51}F/C/B} \\
\midrule
\multirow{4}{*}{Korean} 
& Color & 9.9580 & 5.6810 & 9.8660 & 2.2120 & \textbf{9.9030} \\
& Font & 5.3020 & 9.9300 & 9.7190 & 2.1960 & \textbf{9.9050} \\
& Background & 9.9190 & 9.8830 & 3.7580 & 2.2250 & \textbf{9.9010} \\
& Overall & 8.3930 & 8.4980 & 7.7810 & 2.2110 & \textbf{9.9030} \\
\midrule
\multirow{4}{*}{Arabic} 
& Color & 9.8580 & 4.3420 & 9.9430 & 3.9490 & \textbf{9.3830} \\
& Font & 5.0970 & 9.9540 & 9.3240 & 3.8700 & \textbf{9.4140} \\
& Background & 9.3750 & 9.6770 & 2.8100 & 3.8840 & \textbf{9.4000} \\
& Overall & 8.1100 & 7.9910 & 7.3590 & 3.9010 & \textbf{9.3990} \\
\midrule
\multirow{4}{*}{Japanese} 
& Color & 9.8720 & 5.8450 & 9.5760 & 2.8140 & \textbf{9.7660} \\
& Font & 5.9690 & 9.8920 & 9.5010 & 2.8070 & \textbf{9.7660} \\
& Background & 9.8360 & 9.8770 & 4.3740 & 2.8120 & \textbf{9.7690} \\
& Overall & 8.5590 & 8.5380 & 7.8170 & 2.8110 & \textbf{9.7670} \\
\bottomrule
\end{tabular}
\end{adjustbox}
\caption{Evaluation scores from human raters on synthetic image pairs with controlled variations in text (T), font (F), color (C), and background (B). Each column indicates whether the corresponding attributes are preserved (\ding{51}) or modified (\ding{55}).}
\label{tab:tab_visual_variations_human}
\end{table}

\section{Text Appearance Similarity (TAS)}

\subsection{Disentangled Feature Extraction}

Figure~\ref{fig:tas_examples} presents representative outputs used in the computation of TAS, including examples highlighting variations in text color, font, and background.

\begin{table}[t]
\begin{adjustbox}{width=0.7\columnwidth, center}
\centering
\small
\setlength{\tabcolsep}{7pt}
\begin{tabular}{l|ccc}
\toprule
\textbf{Metric} & \textbf{Korean} & \textbf{Arabic} & \textbf{Japanese} \\
\midrule
SSIM ($\uparrow$) & 0.3244 & 0.3983 & 0.3246 \\
PSNR ($\uparrow$) & 0.5647 & 0.5278 & 0.5124 \\
MSE ($\downarrow$) & $-$0.5647 & $-$0.5278 & $-$0.5124 \\
\midrule
$s_{\text{clr}}$ ($\uparrow$) & 0.5854 & 0.5482 & 0.5323 \\
$s_{\text{fnt}}$ ($\uparrow$) & 0.6805 & 0.5412 & 0.5144 \\
$s_{\text{bg}}$ ($\uparrow$) & 0.7810 & 0.7448 & 0.7857 \\
TAS ($\uparrow$) & \textbf{0.7349} & \textbf{0.6715} & \textbf{0.6582} \\
\bottomrule
\end{tabular}
\end{adjustbox}
\caption{Spearman correlation ($\rho$) between each metric and human evaluation scores across three languages.}
\label{tab:tab_tas_correlation}
\end{table}

\subsection{Metric Validation}

\noindent\textbf{Extended Evaluation on Visual Variations.}
We conduct an experiment to assess whether TAS can more precisely evaluate similarity in color, font, and background compared to conventional image quality metrics such as SSIM, PSNR, MSE, and FID~\cite{fid}. Additional results on synthetic datasets for Arabic and Japanese are provided in Table~\ref{tab:visual_variations_ar_jp}. TAS assigns the highest similarity scores to sets where only the text content differs, whereas other metrics tend to assign the highest scores to sets where font or color is modified, similar to the Korean results.

\noindent\textbf{Correlation with Human Judgment.}
We collected evaluation scores from human raters for each image pair across three languages (Table~\ref{tab:tab_visual_variations_human}). We involved five proficient speakers per language. For each pair, raters independently scored the similarity of text color, font, and background on a scale from 1 (low similarity) to 10 (high similarity). The average of the three scores represents the overall human evaluation.
To assess the reliability of these human judgments, we computed the intraclass correlation coefficient (ICC)~\cite{icc} of type (3, k) across the five raters for each language. The total ICC(3, k) values were 0.9748 for Korean, 0.9473 for Arabic, and 0.9756 for Japanese, indicating excellent inter-rater agreement.
We then computed Spearman correlation coefficients between each TAS component ($s_{\text{clr}}$, $s_{\text{fnt}}$, $s_{\text{bg}}$) and the corresponding human scores, as well as between the overall TAS and the average human ratings. As shown in Table~\ref{tab:tab_tas_correlation}, TAS and its subcomponents exhibit higher correlations with human evaluations compared to conventional image quality metrics (SSIM, PSNR, MSE), demonstrating its effectiveness in assessing visual style similarity.

\begin{figure}[t]
\centering
\includegraphics[width=\columnwidth]{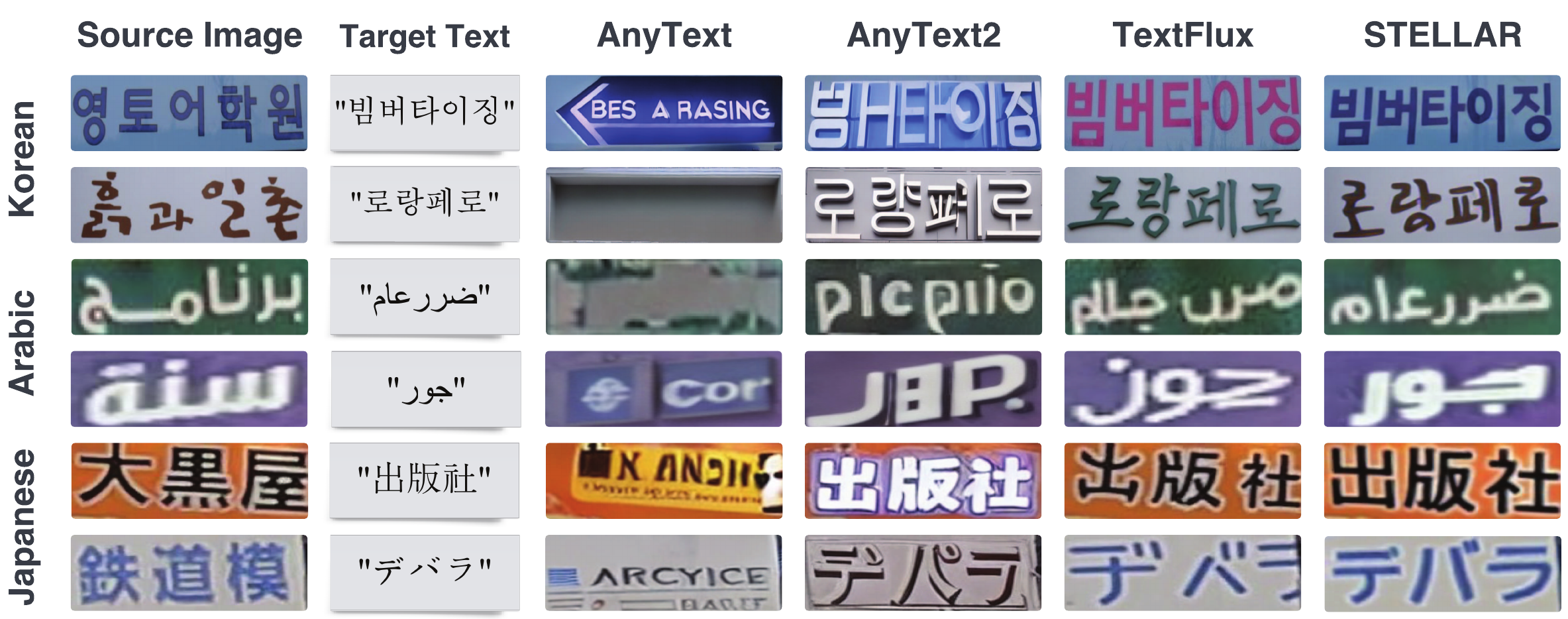}
\caption{Qualitative comparisons of STELLAR and baseline models on three public scene text benchmarks: KAIST STDB (Korean), EvArEST (Arabic), and Billboard JS (Japanese).}
\label{fig:fig_benchmark_qual}
\end{figure}

\begin{table*}[t]
    \begin{adjustbox}{width=0.9\textwidth, center}
    \setlength{\tabcolsep}{2pt}
    \centering
    \small
    \begin{tabular}{l|cccc|cccc|cccc}
    \toprule
    & \multicolumn{4}{c|}{Korean (KAIST STDB)} & \multicolumn{4}{c|}{Arabic (EvArEST)} & \multicolumn{4}{c}{Japanese (Billboard JS)} \\
    \cmidrule(lr){2-5} \cmidrule(lr){6-9} \cmidrule(lr){10-13}
    Metric & AnyText & AnyText2 & TextFlux & \textbf{STELLAR}
           & AnyText & AnyText2 & TextFlux & \textbf{STELLAR}
           & AnyText & AnyText2 & TextFlux & \textbf{STELLAR} \\
    \midrule
    $s_{\text{clr}}$ ($\uparrow$) & 0.6974 & 0.6998 & 0.8027 & \textbf{0.9083}
            & 0.7236 & 0.7519 & \textbf{0.9074} & 0.8842
            & 0.7100 & 0.6749 & \textbf{0.8413} & 0.8370 \\
    $s_{\text{fnt}}$ ($\uparrow$) & 0.6789 & 0.6813 & 0.7503 & \textbf{0.7993}
            & 0.7529 & 0.7441 & \textbf{0.8478} & 0.8389
            & 0.7563 & 0.7411 & 0.8179 & \textbf{0.8224} \\
    $s_{\text{bg}}$ ($\uparrow$) & 0.5702 & 0.3683 & 0.7847 & \textbf{0.8494}
            & 0.5577 & 0.4485 & 0.7452 & \textbf{0.7785}
            & 0.4087 & 0.1223 & 0.5753 & \textbf{0.6263} \\
    TAS ($\uparrow$) & 0.6488 & 0.5832 & 0.7792 & \textbf{0.8523}
            & 0.6781 & 0.6482 & 0.8335 & \textbf{0.8339}
            & 0.6250 & 0.5128 & 0.7448 & \textbf{0.7619} \\
    \midrule
    Rec.Acc ($\uparrow$) & 0.0000 & 0.0869 & 0.5279 & \textbf{0.5590}
            & 0.0000 & 0.0000 & 0.0179 & \textbf{0.3892}
            & 0.0010 & 0.0768 & 0.4029 & \textbf{0.4974} \\
    NED ($\uparrow$) & 0.0218 & 0.2868 & 0.7773 & \textbf{0.8008}
            & 0.0022 & 0.0226 & 0.2671 & \textbf{0.7236}
            & 0.0046 & 0.1305 & 0.6799 & \textbf{0.7917} \\
    \bottomrule
    \end{tabular}
    \end{adjustbox}
    \caption{Quantitative evaluation of STELLAR and baseline models on public scene text datasets (KAIST Scene Text Database, EvArEST Benchmark for Arabic Scene Text, and Billboard in Japanese Streetscapes).}
    \label{tab:tab_other_benchmarks}
\end{table*}

\begin{table*}[t]
\begin{adjustbox}{width=0.8\textwidth, center}
\centering
\small
\setlength{\tabcolsep}{4pt}
\begin{tabular}{l|ccc|ccc|ccc}
\toprule
& \multicolumn{3}{c|}{\textbf{Korean}} & \multicolumn{3}{c|}{\textbf{Arabic}} & \multicolumn{3}{c}{\textbf{Japanese}} \\
\cmidrule(lr){2-4} \cmidrule(lr){5-7} \cmidrule(lr){8-10}
\textbf{Metric} & S1-R & S1-R + S2 & \textbf{STELLAR}
& S1-R & S1-R + S2 & \textbf{STELLAR} 
& S1-R & S1-R + S2 & \textbf{STELLAR} \\
\midrule
SSIM ($\uparrow$) & 0.3778 & 0.4961 & \textbf{0.5061} & 0.3699 & 0.4272 & \textbf{0.4311} & 0.3047 & 0.3218 & \textbf{0.3520} \\
PSNR ($\uparrow$) & 14.6452 & 15.8042 & \textbf{16.1514} & 13.3120 & 13.9098 & \textbf{13.9638} & 13.658 & 13.3676 & \textbf{13.9301} \\
MSE ($\downarrow$) & 0.0437 & 0.0324 & \textbf{0.0301} & 0.0559 & 0.0489 & \textbf{0.0485} & 0.0564 & 0.0580 & \textbf{0.0517} \\
FID ($\downarrow$) & 36.6941 & 35.6152 & \textbf{34.4719} & \textbf{58.9558} & 61.1179 & 62.8025 & \textbf{73.1040} & 80.1315 & 73.7080 \\
\midrule
TAS ($\uparrow$) & 0.8311 & 0.8556 & \textbf{0.8596} & 0.8295 & 0.8607 & \textbf{0.8601} & 0.7549 & 0.7601 & \textbf{0.7714} \\
\midrule
Rec. ACC & 0.6445 & 0.7682 & \textbf{0.8042} & 0.5688 & 0.6769 & \textbf{0.6840} & 0.1532 & 0.2260 & \textbf{0.4338} \\
NED & 0.8274 & 0.8950 & \textbf{0.9115} & 0.7751 & 0.8982 & \textbf{0.8985} & 0.3572 & 0.4522 & \textbf{0.6356} \\
\bottomrule
\end{tabular}
\end{adjustbox}
\caption{
Ablation study on synthetic data filtering in Stage 1. S1-R uses unfiltered synthetic data in Stage 1. S1-R + S2 adds fine-tuning with real-world images in Stage 2. STELLAR employs OCR-filtered synthetic data in Stage 1 and fine-tuning with real-world images in Stage 2.
}
\label{tab:filtering_ablation}
\end{table*}

\begin{figure}[t]
\centering
\includegraphics[width=\columnwidth]{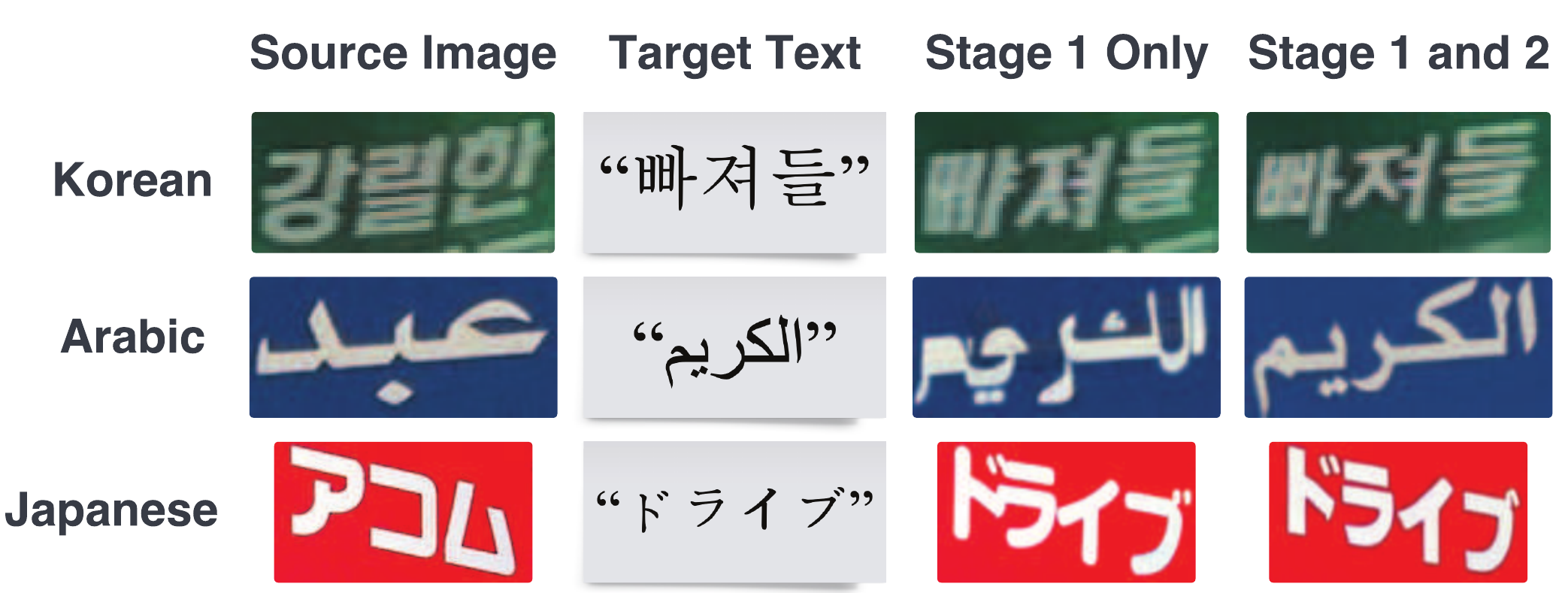}
\caption{Comparison of generation results from models trained only on Stage 1 and those additionally fine-tuned on real-world data in Stage 2.}
\label{fig:qual_stage12}
\end{figure}

\begin{figure*}[t]
\centering
\includegraphics[width=0.92\textwidth]{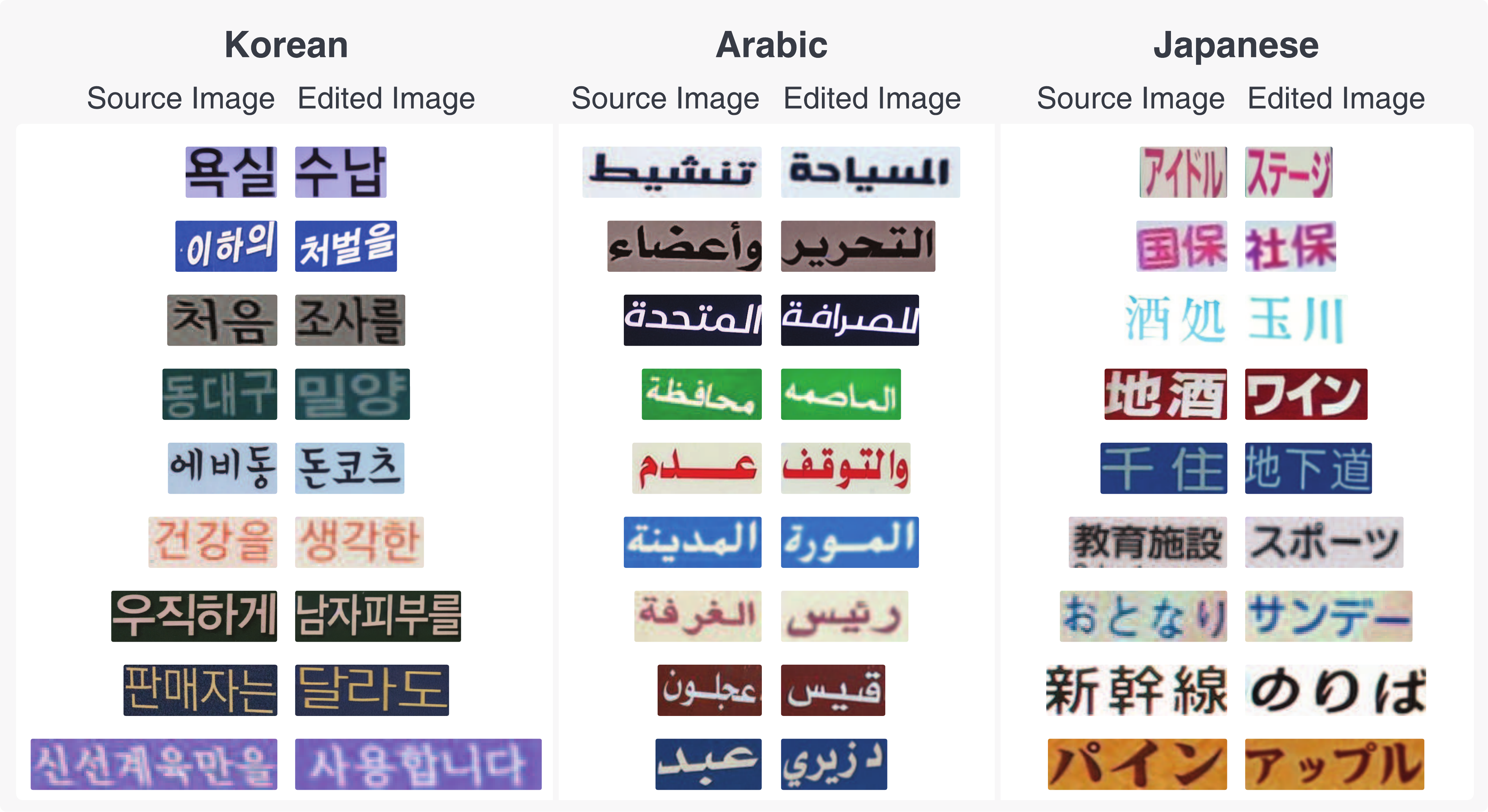}
\caption{Additional examples of source and edited image pairs from STELLAR, demonstrating its ability to edit text content while preserving font, color, and background across Korean, Arabic, and Japanese text images.}
\label{fig:additional_qual}
\end{figure*}

\section{Additional Experiments}

\subsection{Evaluation on Public Benchmarks}

To assess the robustness of STELLAR beyond the STIPLAR evaluation set, we evaluate its performance on three publicly available scene text benchmarks (Table~\ref{tab:tab_other_benchmarks}). The filtering process is identical across all datasets and selects horizontally aligned, single-line, and sufficiently large text regions containing legible characters. For Korean, we adopt the KAIST Scene Text Database~\cite{kaiststdb} (KAIST STDB) and obtain 771 valid samples. For Arabic, we use the EvArEST~\cite{evarest} dataset and extract 948 Arabic text images. For Japanese, we employ the Billboard in Japanese Streetscapes~\cite{billboardjs} (Billboard JS) dataset, where text regions are detected using PPOCRv4, and samples with a recognition confidence above 0.67 are retained, resulting in 963 images for evaluation.

We compute three metrics: TAS, Rec.Acc, and NED on these datasets. STELLAR consistently achieves superior performance compared to baseline models across most languages. While its $s_{\text{clr}}$ and $s_{\text{fnt}}$ scores for Arabic and Japanese are slightly lower than those of TextFlux~\cite{textflux}, the overall TAS and recognition metrics remain higher. Notably, unlike in the STIPLAR evaluation set where STELLAR showed a lower TAS score than TextFlux for Japanese, it achieves a higher score on Billboard JS, indicating improved generalization to unseen real-world data. Representative qualitative examples can be found in Figure~\ref{fig:fig_benchmark_qual}.

\subsection{Impact of Synthetic Dataset Filtering}

To train $G$ in Stage 1, we utilize a synthetic dataset filtered using PPOCRv4 to remove unrecognizable samples. To verify the effectiveness of this filtering process, we conduct an ablation study by removing the filtering step and retraining the model. As shown in Table~\ref{tab:filtering_ablation}, we observe performance degradation in most cases for image quality and style preservation, and consistently for text recognition accuracy. This indicates that constructing a clean training dataset plays a key role in enabling the model to quickly acquire basic text rendering capabilities.

\subsection{Qualitative Comparison}

\noindent\textbf{Effect of Real-world Data Fine-tuning.}
Figure~\ref{fig:qual_stage12} shows qualitative examples comparing outputs from models trained only on Stage 1 versus those trained on both Stage 1 and Stage 2. The former often fails to accurately render text, exhibiting degraded textures, font artifacts, and inability to preserve styles such as outlines. In contrast, the latter generates more accurate and visually consistent results.

\begin{figure}[t]
\centering
\includegraphics[width=\columnwidth]{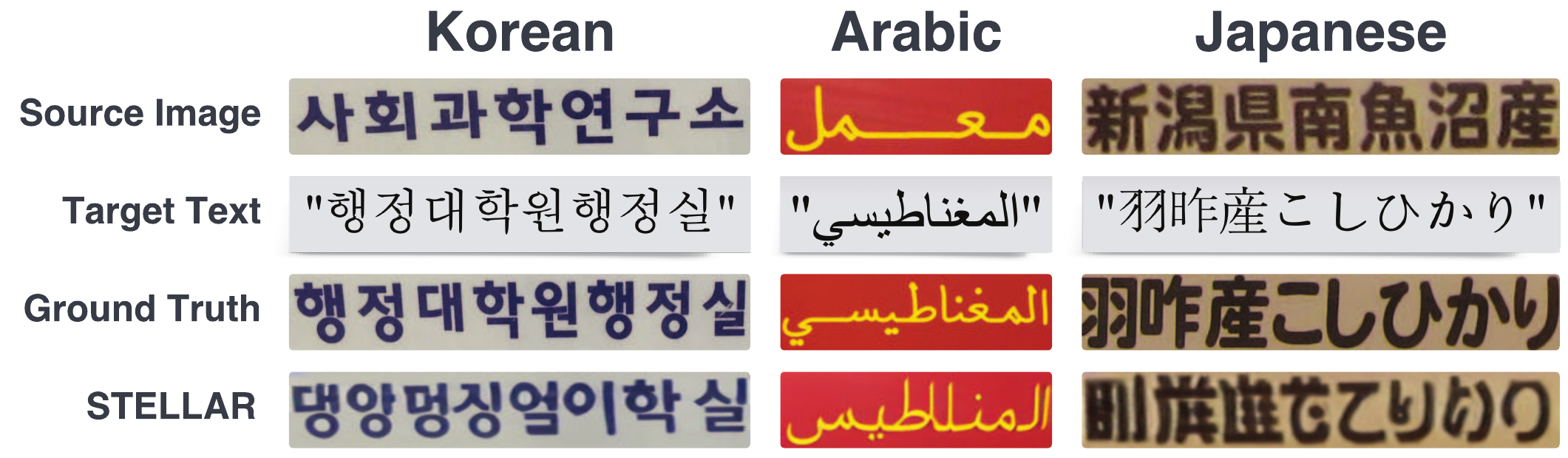}
\caption{Failure cases of STELLAR when editing long text inputs, where the model fails to accurately render the target text.}
\label{fig:fig_limitations}
\end{figure}

\noindent\textbf{Additional Generation Examples.}
Figure~\ref{fig:additional_qual} presents diverse source-generated image pairs from STELLAR, demonstrating its ability to preserve a wide range of text styles and adapt to various real-world scenes.

\noindent\textbf{Limitation on Text Length.}
As illustrated in Figure~\ref{fig:fig_limitations}, STELLAR exhibits noticeable degradation in text rendering quality when processing long text inputs. For target texts exceeding approximately 7–8 characters, the model often generates visually distorted text, indicating reduced reliability in maintaining accurate character rendering. This limitation primarily stems from the scarcity of long-text samples in the training datasets. Collecting and training on a larger number of long-text samples is expected to improve the performance of STELLAR in rendering extended text content.

\end{document}